\documentclass[journal]{IEEEtran}

\usepackage[utf8]{inputenc} % allow utf-8 input
\usepackage[T1]{fontenc}    % use 8-bit T1 fonts
\usepackage{courier}% provides courier font for \texttt
\usepackage[hidelinks,bookmarksdepth=3]{hyperref}  % hyperlinks
\usepackage{url}    % simple URL typesetting
\usepackage{booktabs}  % professional-quality tables
\usepackage{amsfonts}  % blackboard math symbols
\usepackage{nicefrac}  % compact symbols for 1/2, etc.
\usepackage{microtype} % microtypography
\usepackage{lipsum}
\usepackage{graphicx}
\graphicspath{ {./images/} }
\usepackage{amsmath,amssymb,amsfonts}
\usepackage[table]{xcolor}
\usepackage{multirow}

\begin{document}

\title{SSTAF: Spatial-Spectral-Temporal Attention Fusion Transformer for Motor Imagery Classification}

\author{Ummay Maria Muna, Md. Mehedi Hasan Shawon, Md Jobayer, Sumaiya Akter, and Saifur Rahman Sabuj

    \thanks{U. M. Muna, M. M. H. Shawon, M. Jobayer, and S. R. Sabuj are with the Department of Electrical and Electronic Engineering, BRAC University, Dhaka-1212, Bangladesh. (e-mail: maria.muna@bracu.ac.bd, mehedi.shawon@bracu.ac.bd, md.jobayer@bracu.ac.bd and s.r.sabuj@ieee.org)}

    \thanks{S. Akter is with the Department of Electrical and Computer Engineering, University of Maryland College Park, College Park, Maryland 20742, USA. (e-mail: sakter1@umd.edu)}

    \thanks{This work is supported by the Research Seed Grant Initiative-2024, BRAC University, Dhaka-1212, Bangladesh.} 

    \thanks{Corresponding author: S. R. Sabuj (e-mail: s.r.sabuj@ieee.org)}
}

\maketitle

\begin{abstract}
Brain-computer interfaces (BCI) in electroencephalography (EEG)-based motor imagery classification offer promising solutions in neurorehabilitation and assistive technologies by enabling communication between the brain and external devices. However, the non-stationary nature of EEG signals and significant inter-subject variability cause substantial challenges for developing robust cross-subject classification models. This paper introduces a novel Spatial-Spectral-Temporal Attention Fusion (SSTAF) Transformer specifically designed for upper-limb motor imagery classification. Our architecture consists of a spectral transformer and a spatial transformer, followed by a transformer block and a classifier network. Each module is integrated with attention mechanisms that dynamically attend to the most discriminative patterns across multiple domains, such as spectral frequencies, spatial electrode locations, and temporal dynamics. The short-time Fourier transform is incorporated to extract features in the time-frequency domain to make it easier for the model to obtain a better feature distinction. We evaluated our SSTAF Transformer model on two publicly available datasets, the EEGMMIDB dataset, and BCI Competition IV-2a. SSTAF Transformer achieves an accuracy of 76. 83\% and 68. 30\% in the data sets, respectively, outperforms traditional CNN-based architectures and a few existing transformer-based approaches.
\end{abstract}


\IEEEpeerreviewmaketitle

\section{Introduction}
Brain-computer interfaces (BCI) are a revolutionary frontier in neurotechnology that allows researchers to create direct connections between the human brain and external devices without requiring conventional motor output~\cite{lawhern2018eegnet}~\cite{zhang2019making}. Motor imagery (MI) refers to the mental practice or thinking of motor actions or movements of any body part without actual physical movements, as defined by the BCI paradigm. MI-BCIs have shown significant potential in diverse applications, such as neurorehabilitation for stroke patients, assistive solutions for users with motor disabilities, and novel human-computer interaction in the medical sector.

Electroencephalogram (EEG) is widely used in MI-BCI applications due to its non-invasive nature, cost-effectiveness, and sharp temporal resolution~\cite{tang2024spatial}. Despite these advantages, EEG signals inherently present substantial challenges, including temporal variability, low spatial resolution, low signal-to-noise ratio (SNR), significant inter-session and inter-subject variability and the requirement for extensive user training that complicate the decoding of specific motor intentions and limit the widespread development of MI classification tools and techniques~\cite{sun2020eeg}~\cite{lafleur2013quadcopter}~\cite{lotte2014tutorial}. Therefore, developing robust classification algorithms capable of overcoming these limitations might be a critical research direction in advancing MI-BCI technology, specifically for rehabilitation.

Conventional approaches often depend heavily on hand-made feature extraction, which cannot directly extract abstract high-dimensional features from raw data, making it difficult to decode EEG signals. The advent of deep learning has made significant improvements in MI-EEG classification performance~\cite{dose2018end}. Although deep neural networks (DNNs) and convolutional neural networks (CNNs) have shown superior capabilities in automatically extracting high-level spatial features from raw EEG data, these architecture architectures typically focus on local feature extraction and lack the inherent ability to capture long-range dependencies. In recent times, transformer-based architectures have gained attention for their remarkable ability to model long-range dependencies and complex patterns that can be a promising alternative. However, transformers alone often do not fully address the unique multidimensional characteristics of EEG signals, and additional information regarding the spectral and spatial domains would also be required for precise EEG signal decoding.

In this study, we aim to translate brain signals, imagining upper limb movements into commands to automatically control soft robotic gloves for patients with motor dysfunction, creating a reliable neural rehabilitation system. To create an effective neural rehabilitation system, we develop a transformer-based EEG decoding technique called the SSTAF Transformer, fused with an attention mechanism to focus on the most relevant part of the signal. Two widely used publicly available datasets named EEGMMIDB and BCI competition IV 2a were used to illustrate the effectiveness of the proposed model. We demonstrate that multidimensional features are crucial in enhancing the model's performance and increasing the generalizability of the model. Our findings also underline the potential of our methodology to reduce the variability of subjects and the effectiveness of MI-BCIs.

\subsection{Related Work}
Early approaches for MI classification mainly rely on craft-based feature extraction techniques, often combining several strategies with traditional machine learning algorithms. Among these techniques, common spectral features such as band power, power spectral density (PSD), linear discriminant analysis (LDA), and common spatial patterns (CSP) have been widely utilized due to their neurophysiological relevance to motor activities~\cite{blankertz2008berlin}~\cite{jha2013svm}. Although these approaches established important baselines, they often struggled with the high dimensionality and temporal variability of EEG signals, limiting the models' ability to generalize effectively across subjects.

In recent years, many methodologies have been developed and are progressively becoming more advanced. Researchers have shifted from traditional feature engineering to deep learning architectures that automatically learn discriminative representations and have made significant advancements in signal processing and decomposing methods. Schirrmeister et al.~pioneered CNNs, which are among the first deep architectures applied to this domain by introducing the ShallowConvNet and DeepConvNet architectures designed explicitly for EEG decoding that could automatically extract features directly from raw EEG signals and outperformed traditional feature-based approaches~\cite{schirrmeister2017deep}. In subsequent periods, researchers implemented more sophisticated variants incorporating CNNs, including EEGNet by Lawhern et al.~, which utilized depthwise and separable convolutions to create an efficient and compact architecture~\cite{lawhern2018eegnet}. Similarly, Roots et al.~developed EEGNet Fusion using a multi-branch 2D CNN~\cite{roots2020fusion}. Sakhavi et al.~introduced another CNN-based approach that operates on time-frequency representations~\cite{sakhavi2018learning}. At the same time, Zhao et al.~employed a multi-branch CNN architecture to extract complementary features at different time scales~\cite{zhao2019multi} and Tayeb et al.~incorporated hybrid CNN-LSTM architectures to jointly capture spatial patterns and temporal dynamics~\cite{tayeb2019validating}.

CNN-based models often merged with the attention mechanism; for example, the AIDC-CN~\cite{chowdhury2024attention} model included self-attention, cross-attention modules, and capsule networks to extract robust features, and another study integrated spatio-temporal convolution with multi-head attention~\cite{shi2023classification} that shows the effectiveness of combining localized feature extraction that can capture long-range dependencies. Architectures such as the Multi-Branch Fusion CNN~\cite{zhang2023recognition} and MBMANet~\cite{deng2024robust} incorporate parallel pathways to integrate features from different frequency bands and spatial resolutions. Moreover, there are few graph-based methods; for example, SF-TGCN~\cite{tang2024spatial} uses graph convolution to capture complex spatial relationships between EEG channels. Lian et al.~ presented a multitask MI EEG classification neural network by integrating a compact CNN (CCNN), a gated recurrent unit (GRU), and an attention mechanism for dynamic feature fusion~\cite{lian2024end}, while Kumari et al.~proposed another novel method using MRMR for optimal channel selection and a hybrid WSO-ChOA for feature selection. A two-tier deep learning framework integrates CNN for temporal feature extraction and M-DNN for spatial analysis~\cite{kumari2024eeg}. PMD-MSNet~\cite{dai2025periodicity} by Dai et al.~transformed 1D EEG signals into multiperiod 2D vectors to extract periodic and spatiotemporal features, while a framework like AM-EEGNet~\cite{lin2024eegnet} was developed for a specific application, namely to classify the EEG states of stroke patients. All of these methods perform better than conventional ConvNets.

Transformer architectures have recently been adapted for EEG signal analysis~\cite{abibullaev2023deep} due to their ability to capture long-range dependencies in sequential data, although initially developed for natural language processing~\cite{vaswani2017attention}. Kostas et al.~introduced a transformer-based approach called BENDR for general EEG representation learning that performed impressively on various neurological classification tasks~\cite{kostas2021bendr}. Song et al.~proposed EEG-Conformer with an attention-based transformer applied to capture temporal dependencies in EEG signals~\cite{song2022eeg} and Liao et al.~developed EEGEncoder~\cite{liao2024eegencoder}, leveraging a dual-stream temporal-spatial block (DSTS) combined with temporal convolutional networks (TCN) for local feature extraction and transformer layers for capturing global channel dependencies. Similarly, Zhao et al.~introduced CTNet~\cite{Zhao2024ctnet} and Zhang et al.~introduced local and global convolutional transformer-based model~\cite{ zhang2023local}. Furthermore, EEG-VTTCNet~\cite{shi2024eeg} employed a joint loss training strategy that balances the features of a vision transformer (ViT) branch and a TCN branch. These approaches achieved higher classification accuracies on benchmark datasets compared to previous studies, proving the benefits of transformer-based methods in managing non-stationary EEG signals.

\begin{figure*}[!ht]
    \centering
    \includegraphics[width=\textwidth]{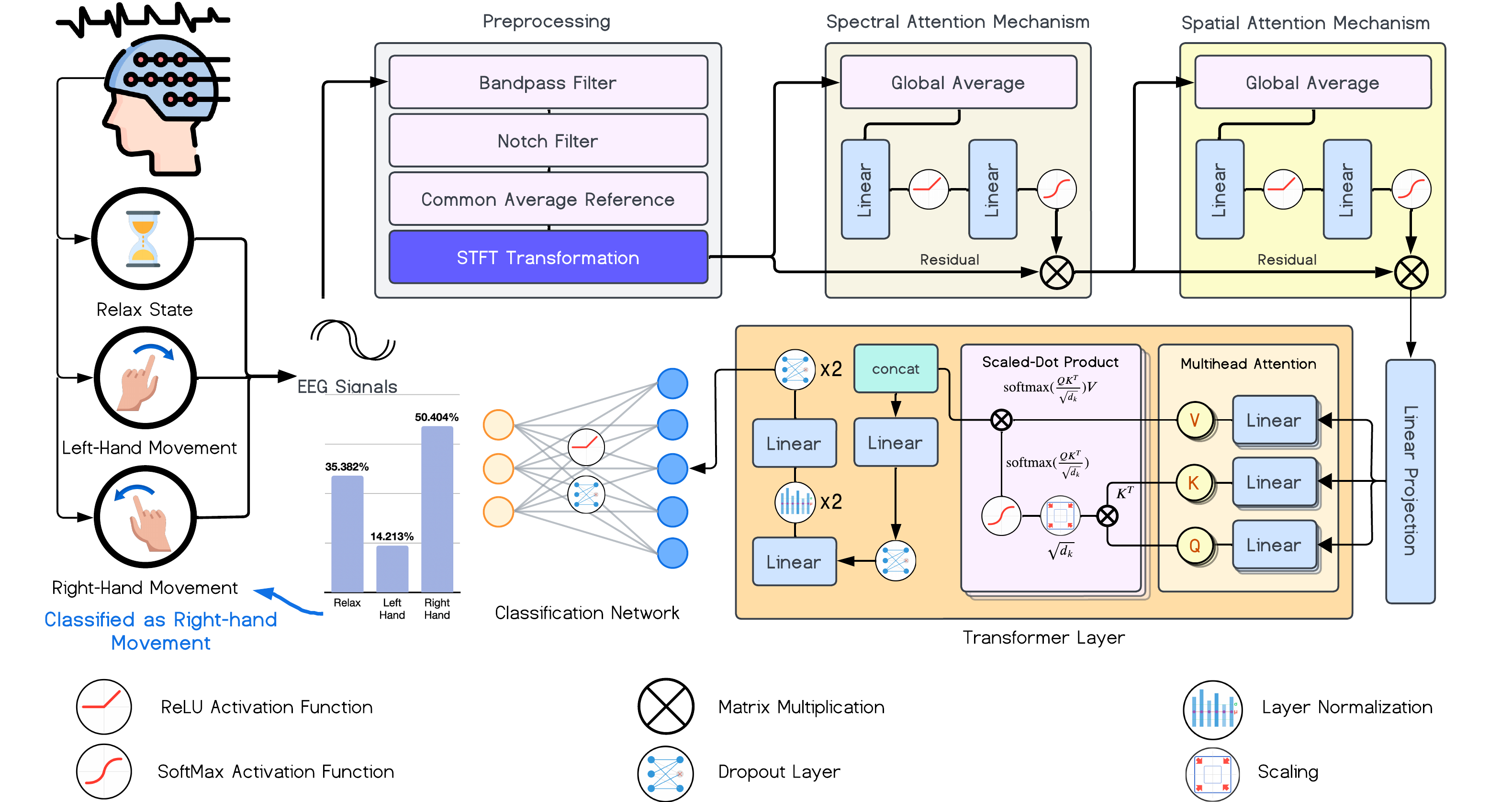}
    \caption{\textbf{Detailed Model Architecture}, The proposed SSTAF Transformer model for EEG MI classification. Continuous signal data is preprocessed, followed by extracting spectral and spatial information with their corresponding attention module and lastly goes through the transformer layer(s). A classifier network calculates the probability of the class label.
    \newline
    \label{fig:expanded_model}}      
\end{figure*}

\subsection{Contribution and Organization}
A persistence challenge in MI-BCI research is achieving robust cross-subject generalization. Palazzo et al.~highlighted that performance significantly degrades when models are trained on specific subjects and tested on unseen subjects~\cite{palazzo2020correct}. Several approaches have been proposed to address this problem, such as transfer learning techniques~\cite{wu2020transfer}, domain adaptation methods~\cite{ozdenizci2020learning}, and data augmentation strategies~\cite{patil2024coordconformer}. This study addresses the same problem by creating standardized spectral-temporal representations emphasizing common neural patterns while reducing subject-specific artifacts.

This paper investigates improving the accuracy of cross-subject MI EEG classification for BCI applications. In contrast to existing approaches that process EEG signals in isolated domains, there is a research gap in fully leveraging multi-domain transformer architectures with attention mechanisms for EEG-based MI classification. The main contributions of this paper include:  

\begin{enumerate}
\item We propose a novel transformer-based framework integrated with spectral-spatial-temporal attention fusion techniques named SSTAF Transformer. 
\item We designed a custom multi-channel STFT transformation that efficiently EEG signals into a structured 4D time-frequency representations to facilitate subsequent attention mechanisms.

\item We create a specialized attention fusion mechanism that helps the model to dynamically attend to the most discriminative spectral-spatial-temporal patterns and reduce subject variations while remaining consistent across subjects.

\item Finally, extensive experiments are carried out using the EEGMMIDB and BCI Competition IV 2a datasets with leave-one-subject-out and K-fold cross-validation to validate the effectiveness of our proposed SSTAF Transformer.
\end{enumerate}

The remainder of the paper is organized as follows: Section II analyzes the system model and data preprocessing, Section III discusses the proposed model architecture, Section IV presents the results and discusses them, and Section V concludes the study.

\section{System Model and Data Preprocessing}
\subsection{System Scenario}
In the EEGMMIDB dataset, EEG signal data related to MI hand movements were collected using EEG electrodes placed on the sensorimotor cortex of the scalp, where participants were instructed to imagine specific movements without performing physical movements. The users focused on a computer screen that displayed directional targets during each trial and performed MI corresponding to the target's location. After acquiring EEG signals, extensive preprocessing techniques are applied, such as bandpass filter, notch filter, and common average reference. Subsequently, the time-frequency domain features are extracted from the preprocessed data, which are passed to the model, starting from the spectral attention block and the spatial attention block sequentially to get the features across the spectral and spatial domains. Through a linear projection, the features enter into a transformer module consisting of multi-head attention, which helps in deriving long-range temporal relationships and complex relationships among the signals. Lastly, a couple of linear layers work as a final classifier between the signals being a relaxed state, a right-hand movement, or a left-hand movement. Figure~\ref{fig:expanded_model} shows the detailed architecture of the proposed model.

\begin{table*}[!ht]
\centering
\caption{Summary of The EEG motor imagery datasets.\label{tab:EEG_datasets}}   
    \begin{tabular}{l c c c c c c c c}
\toprule
Dataset Name & Subjects & Channels & Sampling Rate & MI Tasks & Type & Total Trials & Duration \\
\midrule
EEGMMIDB & 103 & 64 & 160 Hz & Left/right hand & Imagined & 180 & 4 sec \\
BCI IV-2a & 9 & 22 & 250 Hz & Left/right hand, foot & Imagined & 288 & 4 sec \\
\bottomrule \\
    \end{tabular} 
\end{table*}

\subsection{Dataset and Preprocessing}
This study uses the EEGMMIDB as the primary dataset, which is a curated version of the PhysioNet EEG Motor Movement/Imagery Dataset \cite{shuqfa2024physionet, goldberger2000physiobank}. In the original dataset recording session, 109 healthy subjects participated in recording EEG signals. Six subjects are excluded due to their noise and invalidity. They recorded the signals with a 160 Hz sampling rate using the BCI2000 system \cite{schalk2004bci2000}, which has 64 electrode channels according to the international 10-10 system. Each subject performed 12 runs of four distinct tasks, including two motor execution (ME) and two motor imagery (MI). Motor execution tasks involve opening and closing either of the left or right fists (unilateral) or both fists (bilateral), while MI tasks involve the same actions but in imagination without any physical movement. There are three runs for each type of task. Each run consisted of approximately 30 trials preceded by a short relaxation period. The dataset provides both raw EEG signals and the corresponding annotations for each test, indicating the type and timing of the task. In this study, we focus specifically on MI tasks related to hand movements. To demonstrate the robustness of our model, we also experimented with the BCI Competition IV 2a dataset \cite{brunner2008bci}. Brief information about both datasets is demonstrated in Table ~\ref{tab:EEG_datasets}.

\subsection{Data Filtering and Enhancement}
According to the EEGMMIDB dataset, tasks 1 and 3 contain motor execution (ME) activity, and tasks 2 and 4 contain Motor Imagery (MI) activity. After excluding ME tasks 1 and 3 from the dataset, we are left with only MI activities, represented by tasks 2 and 4. Each task is performed in various runs; for the MI activities, the run numbers are \texttt{02, 04, 06, 08, 10, 12}. Each run has its own set of annotations, and in total, the possible annotations for MI activities are  \texttt{4, 5, 6, 10, 11} and \texttt{12}. After exclusion, with six runs left for each subject, a total of 618 runs have been preserved, which represents only MI activities. The corresponding EEG signals and their annotations are preserved as paired data structures for each retained run to maintain data integrity throughout subsequent processing steps.
 
After filtering the dataset, the EEG signals are preprocessed to enhance signal quality. The preprocessing steps included bandpass filtering, notch filtering, and common average reference (CAR) application. In EEG signal processing, a common preprocessing pipeline involves several steps to remove unwanted noise and artifacts, starting with a bandpass filter mechanism designed with a bandpass of 8 Hz to 30 Hz to capture the frequency range $\mu$ and $\beta$ sensorimotor rhythms, which are commonly associated with MI tasks~\cite{yu2022effects}. The bandpass filter is applied to each EEG signal channel individually. Mathematically, if the signal function is denoted by $\varphi(\cdot)$; where $c$ is the channel number, and $t$ is the time index, then the bandpass filtered signal $X^{bf}$ can be obtained from the equation~\ref{eq:band_pass}:
\begin{equation}
    X^{bf}_{c, t} = \mathcal{B}(\varphi(c, t)),
    \label{eq:band_pass}
\end{equation}
where $\mathcal{B}(\cdot)$ is the bandpass filter function, $X^{bf}_{c, t}$ is the bandpass filtered signal. For the bandpass filter function, the Butterworth filter is applied to each signal on the order of 5~\cite{butterworth1930theory}. Following this, a notch filter, $\mathcal{N}$ is applied to each bandpass filtered EEG signal channel individually to remove the 50 Hz powerline noise \cite{leske2019reducing}. Equation~\eqref{eq:notch_filter} represents the notch filter function.
\begin{equation}
    X^{nf}_{c,t} = \mathcal{N}(X^{filtered}_{c,t}).
    \label{eq:notch_filter}
\end{equation}

Finally, the common average reference (CAR)~\cite{ludwig2009using} is utilized to identify the actual signal sources in very noisy recordings and remove this common noise by subtracting the average across all channels from each individual channel of the notched signal, computed as equation~\eqref{eq:CAR}:
\begin{equation}
    X^{CAR}_{c,t} = X^{notched}_{c,t} - \frac{1}{N}\sum_{c=1}^{N} X^{notched}_{c,t},
    \label{eq:CAR}
\end{equation}
where $X^{CAR}_{c,t}$ is the CAR signal, $X^{notched}_{c,t}$ is the notched signal, and $N$ is the total number of channels. Together, by effectively reducing noise and focusing on the neural signals of interest, followed by powerline interference and subtracting the average signal, it enhances the clarity and accuracy of the EEG signal, allowing more reliable detection and interpretation of neural activity related to MI tasks.
\begin{figure}[!h]
    \centering
    \includegraphics[width=0.48\textwidth]{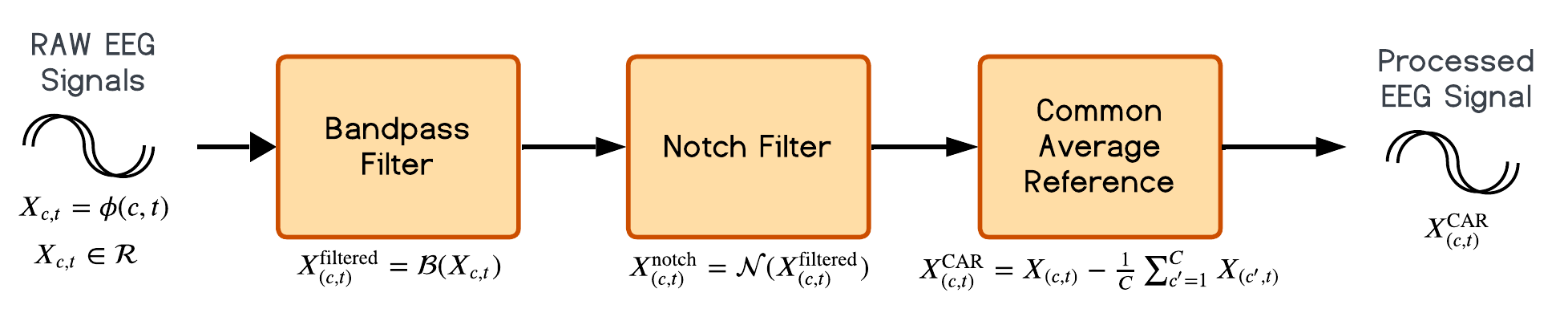}
    \caption{\textbf{Data Preprocessing}, Bandpass Filter, Notch Filter, and Common Average Reference are applied to the raw signal data. \label{fig:data_preprocessing_1}}
\end{figure}

\subsection{Epoch Extraction}
\begin{figure*}[!h]
    \centering
    \includegraphics[width=\textwidth]{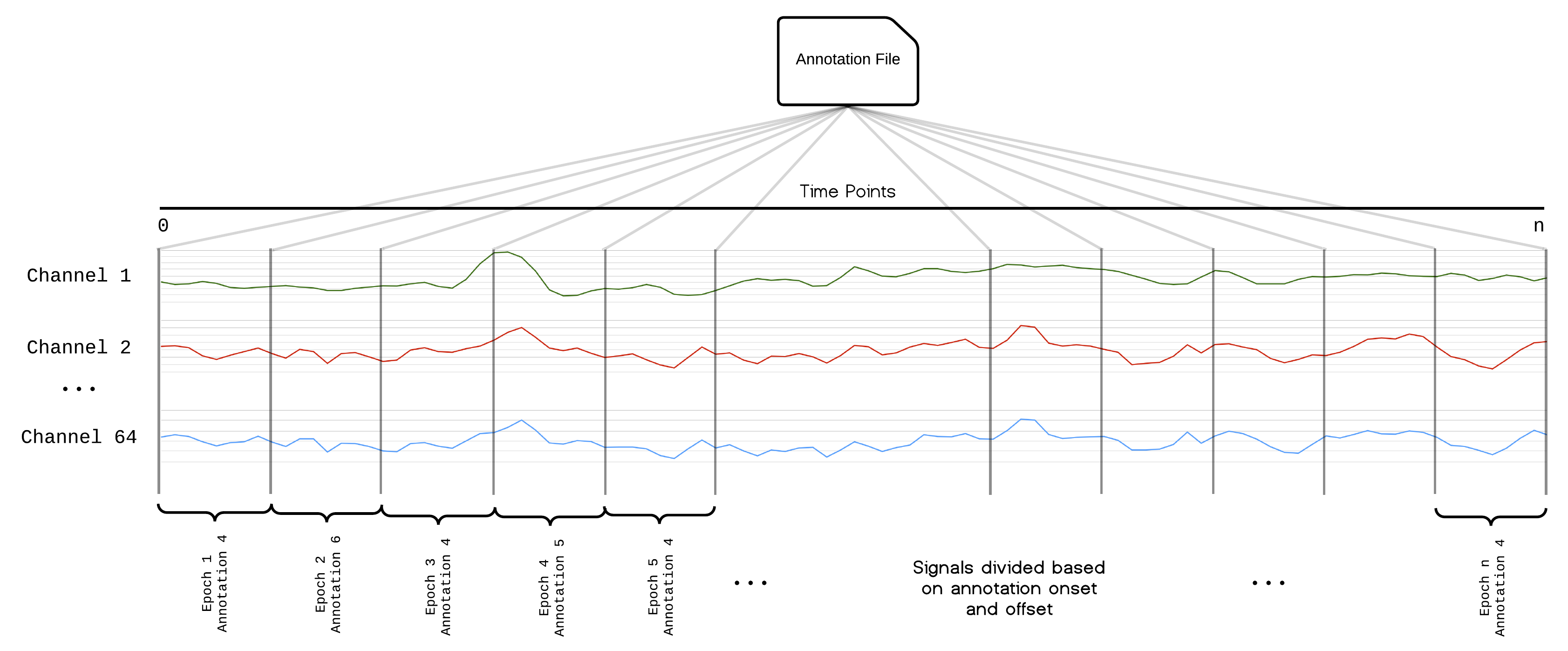}
    \caption{\textbf{Epoch segmentation}, Each signal file are converted into epochs based on the annotation provided in the annotation file. Each signal is segmented into 30 segment. Each segment has their own annotation and 64 channel data with time points. \label{fig:epoch_extraction}}
\end{figure*}

The EEG signals are segmented into epochs based on the corresponding annotation files. Each EEG signal had 30 trials and each trial is preceded by a short relaxation period. The EEG signals are segmented into epochs based on the start and end times of each trial from the annotation file. Our study excluded foot tasks, as it focuses on unilateral operation of the upper extremity on MI tasks. With the 618 runs and 30 trials per run, a total of 18,540 epochs are extracted. Approximately 14 thousand segments are retained after removing the unwanted annotations. Figure~\ref{fig:epoch_extraction} visualizes the epoch segmentation process. Segmenting continuous EEG data into epochs and removing irrelevant annotations helps focus the analysis on the most meaningful time segments related to MI.

After segmentation, the epochs had varying lengths depending on the duration of the trials. Considering the lengths, the epochs are cropped, padded, or totally discarded. This is to ensure that all epochs have a uniform length, which is essential for consistent feature extraction and deep learning model training. The epoch lengths are standardized to a fixed length of 640 data points or 4 seconds duration, 160 data points per sample. The annotation lengths and the respective actions against them are listed below.
\begin{itemize}
    \item Shape (656, 64) has 11844 epochs: \textsc{sliced to shape (640, 64)}
    \item Shape (672, 64) has 1944 epochs: \textsc{sliced to shape (640, 64)}
    \item Shape (688, 64) has 18 epochs: \textsc{sliced to shape (640, 64)}
    \item Shape (640, 64) has 25 epochs: \textsc{kept as it is}
    \item Shape (416, 64) has 63 epochs: \textsc{discarded because of the short length}
    \item Shape (624, 64) has 11 epochs: \textsc{discarded because of the short length}
\end{itemize}

Annotation filtering resulted in exactly 13,831 epochs with a fixed shape of 640 time points with 64 channels. Then we apply standardization on this prepared dataset, since a standardized epoch length ensures a uniform structure among all data segments, important for consistent feature extraction and analysis, especially with deep learning models.

\subsection{Data Standardization}
Standardization represents a critical step in the preprocessing of EEG data to improve the performance of deep learning algorithms. We used the z-score normalization technique that transforms EEG signals from all segments so that each channel exhibits zero mean and unit variance throughout the entire dataset. Given an EEG segment $X \in \mathbb{R}^{T \times C}$, where $T$ represents the temporal samples and $C$ denotes the number of channels, we apply standardization as follows:
\begin{equation}
    X_{\text{standardized}} = \frac{X - \mu}{\sigma + \epsilon},
\end{equation}
where $\mu \in \mathbb{R}^C$ represents the mean value calculated across all temporal points and training samples for each channel, $\sigma \in \mathbb{R}^C$ denotes the corresponding standard deviation, and $\epsilon$ (set to $10^{-8}$) is a very small constant added to halt division by zero. This standardization approach improves the stability of gradient-based optimization and allows the subsequent deep learning model to focus on relevant discriminative patterns rather than amplitude variations between recording sessions or subjects.

\subsection{Short-Time Fourier Transform}
The STFT is applied to extract time-frequency representations from EEG signals. The STFT transforms a time-domain EEG signal into the frequency domain for short, overlapping segments, providing both temporal and spectral information simultaneously. Given an EEG epoch signal of length L for each channel, the STFT, \(X[m,k]\) is defined as:
\begin{equation}
X[m, k] = \sum_{n=0}^{L-1} x[n + m R] w[n] e^{-j \frac{2\pi}{L} k n},~~k = 0, 1, \dots, L-1,
\end{equation}
where \(x[n+mR]\) means that original time-domain signal \(x[n]\) but shifted by \(mR\) to represent the part of the signal in the \(m_{th}\) window, \(w[n]\) is the window function, \(n\) is the index within each window, \(L\) is the window length (also called \( n_{fft}\), \(m\) denotes the index of the windowed segment, \(k\) refers to the frequency bin index, and \(R\) is the hop size (stride) between consecutive segments. For each EEG epoch (batch) and each channel individually, the EEG signal \(x[n]\) is processed segment by segment. The procedure involved the following steps:
\begin{itemize}
    \item The EEG signal \( x[n] \) is divided into overlapping segments using a Hann window with length \( n_{fft}\) and hop size \(R\).
    \item The FFT is computed for each window segment, resulting in complex STFT coefficients \( X[m, k] \).
    \item The magnitude of these complex STFT coefficients is computed to obtain the power spectrum \( P[m, k] = |X[m, k]|^2 \).
\end{itemize}

\begin{figure}[!h]
    \centering
    \includegraphics[width=0.48\textwidth]{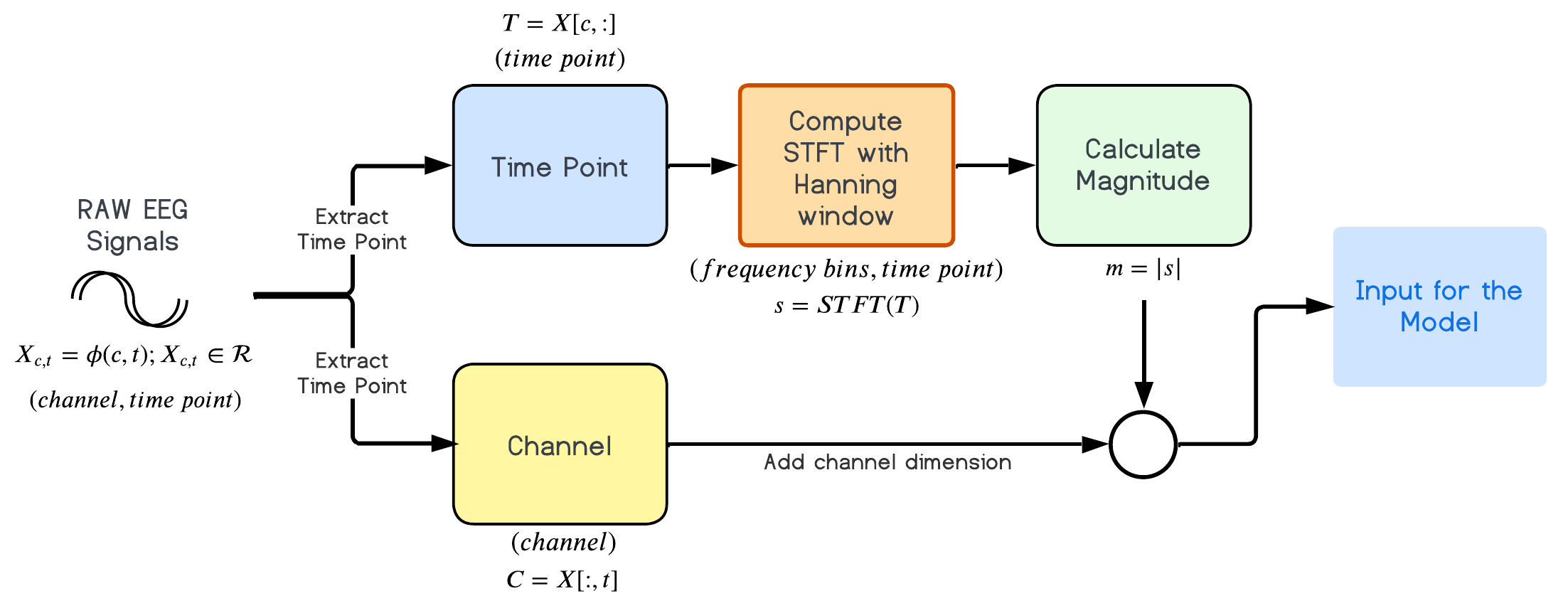}
    \caption{\textbf{Short Time Fourier Transform}, mechanism of STFT for extracting time-frequency domain features from EEG signals. \label{fig:short-time-fourier-transform}}
\end{figure}

After computing the power spectra for each EEG channel separately, the resulting power spectra are combined, forming tensors with dimensions \((f_b, t_f)\), where \( f_b \) is the number of frequency bins and \( t_f \) is the number of time frames. These tensors are stacked in all EEG epochs to produce the final features of shape \((n_{channels}, f_b, t_f)\).

In this study, the sampling frequency \( fs = 160 \) Hz, the window length \( n_{fft} = 128 \), and the hop length \( H_l = 64 \). This approach captures localized temporal and spectral features within EEG signals and enables the effective extraction of dynamic features relevant to MI classification tasks. Figure~\ref{fig:short-time-fourier-transform} shows the time-frequency feature extraction process from EEG data.

\section{Proposed Model Architecture}
The proposed model consists of a spectral attention module and a spatial attention module, followed by a transformer block and a classifier network.

\subsubsection{Spectral Attention Module}
The spectral attention module is designed to extract spectral or frequency domain features from EEG signals using an attention mechanism. This module takes the transformed STFT feature vector as input and processes them through a multi-layer perceptron which calculates the attention weights with a single attention head. The module takes an input vector $\boldsymbol{x}$ with shape $(b, c, f_{bin}, t)$, where $b$ denotes the batch size, $c$ is the number of channels, $f_{bin}$ is the number of frequency bins, and $t$ is the number of time frames. We defined the frequency bin ($f_{bin}$) as 65. Before passing the input to the attention mechanism ($AM_{spectral}$) which is an MLP, it computes the mean across the channels and time frame dimensions, is shown in equation~\eqref{eq:spectral}. 
\begin{equation}
\mathbf{AM}_{\text{spectral}}(\boldsymbol{x}) = \mathrm{Softmax}\left( \mathbf{W}_2 \cdot \mathrm{ReLU}\left( \mathbf{W}_1 \boldsymbol{x} + \boldsymbol{b}_1 \right) + \boldsymbol{b}_2 \right),
\label{eq:spectral}
\end{equation}
where $\mathrm{ReLU()}$ and $\mathrm{Softmax()}$ are the activation functions. In equation~\eqref{eq:spectral2}, the derived mean tensor $\mathbf{X}_{\text{mean}}$ of shape $(b, f_b)$ serves as a global descriptor for each frequency bin, capturing spectral information across all channels and time frames.
\begin{equation}
\mathbf{X}_{\text{mean}} = \frac{1}{C} \sum_{c=1}^{C} \mathbf{X}_{c, f_{\text{bin}}, t},
\label{eq:spectral2}
\end{equation}

\begin{figure}[!h]
    \centering
    \includegraphics[width=0.48\textwidth]{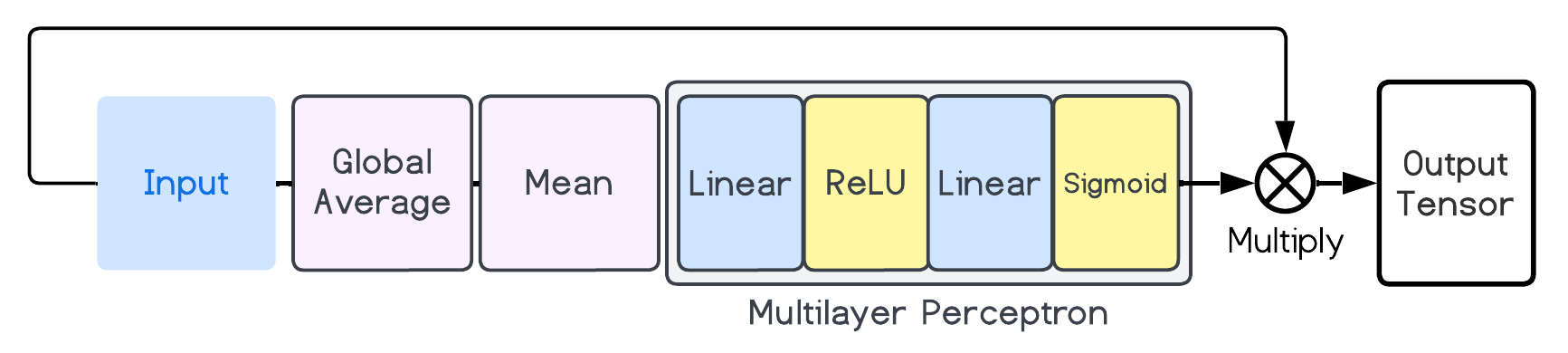}
    \caption{\textbf{Single headed Attention Module Architecture}, which is used both in spectral and spatial attention architecture.\newline
    \label{fig:sp_sp_architecture}}
\end{figure}

The attention mechanism is applied to the $\mathbf{X}_{\text{mean}}$ tensor to calculate the attention weights by passing the tensor through a combination of linear layers with activation functions $\mathrm{ReLU}$ and $\mathrm{Softmax}$. The output of the attention layer is multiplied by the input tensor $\mathbf{X}$ to get the final output tensor $\mathbf{X}_{\text{spectral}}$, which is then passed to the next module. The attention mechanism can be represented as the equation~\eqref{eq:spectral3}.
\begin{equation}
\mathbf{X}_{\text{spectral}} = \mathbf{X} \cdot \mathbf{AM}_{\text{spectral}}\left(\mathbf{X}_{\text{mean}}\right),
\label{eq:spectral3}
\end{equation}
The output of this module is a tensor of attention scores of shape ($b$, $f_b$) which is then multiplied with the input tensor to obtain the final output tensor $\mathbf{X}_{\text{spectral}}$ with shape ($b$, $c$, $f_{bin}$, $t$). Figure~\ref{fig:sp_sp_architecture} shows the architecture of the single-headed attention module.

\subsubsection{Spatial Attention Module}
Similar to the spectral attention module, the spatial attention module receives the output of the previous module as input. This module processes this input tensor $\boldsymbol{x}$ through a multi-layer perceptron to calculate the attention weights, but in this module, the output of the MLP represents the channel descriptor. Another difference is that this module computes the mean across the frequency bins and time frames dimensions, instead of the channels and time frames. Figure~\ref{fig:sp_sp_architecture} shows the architecture of the single headed attention module used in this spatial attention mechanism.

This module takes the output of the Spectral Transformer module as input and calculates the attention score. The Spectral module is applied first because it boosts the important frequencies in the signal and then focuses on the channels that carry the most useful information. By first adjusting the frequency details, it is ensured that the subsequent channel adjustment works on the already refined signal. Considering this, the two modules (Spectral and Spatial) can be thought of as a one module with two stages, which applies attention to the signal in two different dimensions inclusively. Similar to the equation~\eqref{eq:spectral}, the spatial attention can be calculated using equation~\eqref{eq:spatial}:
\begin{equation}
\mathbf{AM}_{\text{spatial}}(\boldsymbol{x}) = \mathrm{Softmax}\left( \mathbf{W}_4 \cdot \mathrm{ReLU}\left( \mathbf{W}_3 \boldsymbol{x} + \boldsymbol{b}_3 \right) + \boldsymbol{b}_4 \right),
\label{eq:spatial}
\end{equation}

The mean is also computed according to the equation~\eqref{eq:spectral2} which captures the spatial information across all frequency bins and time frames. Finally, the attention mechanism is applied to the $\mathbf{X}_{\text{mean}}$ tensor to calculate the attention weights that to be multiplied by the input tensor $X$ to obtain the final output tensor $\mathbf{X}_{\text{spatial}}$ with shape ($b$, $c$, $f_{bin}$, $t$). The frequency bin ($f_{bin}$) for the spectral transformer is also set to 65. The attention mechanism for the spatial domain can be calculated as equation~\eqref{eq:spatial2} :
\begin{equation}
\mathbf{X}_{\text{spatial}} = \mathbf{X} \cdot \mathbf{AM}_{\text{spatial}}\left(\mathbf{X}_{\text{mean}}\right),
\label{eq:spatial2}
\end{equation}

Finally, the outputs from both the spectral and spatial transformer modules are multiplied to form an input tensor that is also an informative spatio-spectral representation for the subsequent transformer block.

\subsubsection{Transformer Module}
The Transformer based on the study `Attention is all you need'~\cite{vaswani2023attentionneed} which is widely used to process sequential data. As EEG signals are inherently sequential time series data, transformer-based approaches are particularly suitable for processing them. The transformer module works as one of the key components of the entire architecture. It comprises $L_n$ layers of the multi-head self-attention mechanism with $h_n$ heads, incorporating multiple feed-forward neural networks.

It is a $N$ layered encoder-decoder architecture with stacked multi-head attention (MHA) mechanism followed by fully connected feed forward layers, dropout layers, and layer normalization submodules. In this research, the transformer is modified to two layers in depth and four attention heads. The attention head is composed of the matrices, query $\mathbf{Q}_{\text{i}}$, key $\mathbf{K}_{\text{i}}$, and value $\mathbf{V}_{\text{i}}$ that are used to compute the attention scores, focusing on four different parts of the input sequence. The matrices are computed as follows:
\begin{equation}
\mathbf{Q}_{\text{i}} = \mathbf{X} \mathbf{W}^Q_{\text{i}}, \quad 
\mathbf{K}_{\text{i}} = \mathbf{X} \mathbf{W}^K_{\text{i}}, \quad 
\mathbf{V}_{\text{i}} = \mathbf{X} \mathbf{W}^V_{\text{i}},
\end{equation}
where $\mathbf{X}$ is the input tensor; and $\mathbf{W}^Q_{\text{i}}$, $\mathbf{W}^K_{\text{i}}$, and $\mathbf{W}^V_{\text{i}}$ are the weight matrices for the query, key, and value, respectively. Attention scores are calculated by applying the scaled dot product in $\mathbf{Q}_{\text{i}}$, $\mathbf{K}_{\text{i}}$ and $\mathbf{V}_{\text{i}}$, as shown in equation~\eqref{eq:attention_score}:

\begin{equation}
\mathbf{Z}_{\text{i}} = \mathrm{Attention}(\mathbf{Q}_{\text{i}}, \mathbf{K}_{\text{i}}, \mathbf{V}_{\text{i}}) = \mathrm{softmax}\left( \frac{\mathbf{Q}_{\text{i}} \mathbf{K}_{\text{i}}^{\mathsf{T}}}{\sqrt{d_k}} \right) \mathbf{V}_{\text{i}},
\label{eq:attention_score}
\end{equation}
where $\mathbf{Z}_{\text{i}}$ is the output of the attention mechanism for the $i^{th}$ head and $d_k$ is the key vector dimensions, which scales the dot product to prevent large values that can lead to saturation of the $\mathrm{Softmax}$ function. Subsequently, the attention scores are concatenated to create the final output of the multi-head attention mechanism, which is computed as equation~\eqref{eq:final_attention}:
\begin{equation}
\mathbf{Z} = \sum_{i=1}^{h} \mathbf{Z}_{\text{i}} \mathbf{W}_{\text{i}}^O,
\label{eq:final_attention}
\end{equation}
where $\mathbf{Z}$ is the final output of the multi-head attention module, $h$ is the number of attention heads, $\mathbf{Z}_i$ is the output of the $i^{th}$ attention head, and $\mathbf{W}_i^O$ is a specific learnable output projection matrix for head $i$ that is used to project each output of the attention head into a common space. Lastly, the output of the multi-head attention mechanism is passed through multiple feed-forward neural networks where the dropout layer and layer normalization are used in between, as shown in Figure~\ref{fig:expanded_model}.

\begin{figure}[!h]
    \centering
    \includegraphics[width=0.48\textwidth]{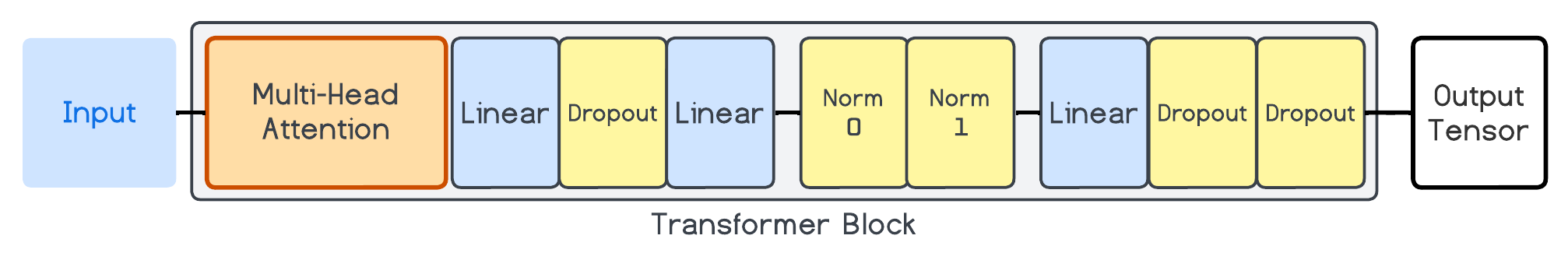}
    \caption{\textbf{Simplified transformer block}, as defined in the PyTorch library, used in the proposed model architecture with some parameter modifications. \label{fig:transformer_block}}
\end{figure}

Prior to feeding the data to the transformer encoder shown in Figure~\ref{fig:transformer_block}, the input is projected onto a fixed dimension of $(b, t, d_h)$ where $b$ represents the batch size, $t$ denotes the number of time frames and $d_h$ is the hidden dimension. The projected data are passed through the transformer block as shown in Figure~\ref{fig:expanded_model}. The transformer encoder is composed of two identical layers, each containing a multi-head self-attention mechanism followed by a fully connected feed-forward neural network. The multi-head attention includes $n_{head}=8$ parallel attention heads, the hidden dimension of $d_{hidden}=64$, and the feed-forward dimension $d_{ff}=256$, is set by quadrupling the hidden dimension. Moreover, the Gaussian error linear unit (GELU) works as the activation function within these transformer layers. Figure~\ref{fig:transformer_block} shows a simplified architecture of the transformer block.

\subsubsection{Classifier Network}
The transformer encoder generates a sequence of feature vectors across time dimensions. To create a single representation for the entire trial, these time-based features are averaged together. This mean value condenses all the information into one compact feature vector that is then used for classification. The feature vector is then passed through a classifier network consisting of two linear layers to predict the final class labels. The first layer reduces the feature dimensionality by half, followed by an activation function ReLU and a dropout layer. The final linear layer maps the features to the output space corresponding to the number of classes in the MI task for a given batch. This simple network efficiently translates the learned spectral, spatial, and temporal representations into class predictions.

\section{Results and Experiments}

\subsection{Implementation Details}
The proposed SSTAF Transformer model is implemented using the PyTorch deep learning framework, and train tests are performed with CUDA support on both local machine and cloud servers. To prove the effectiveness of the proposed model, the model is rigorously evaluated across two widely used publicly available datasets, one is EEGMMIDB and another is BCI Competition IV-2a. These two datasets consist of 103 and 9 subjects, respectively, and the study focuses on distinguishing brain signals while resting and imagining movements of the left/right hand. AdamW optimizer is used to train the model, the learning rate ranges from 0.001 to 0.0001, and the batch size is 16. In addition, the dropout value is set to 0.1 for the spectral and spatial attention module and 0.2 for the transformer module. The performance of the model was evaluated at different epochs, and 20 was found as an optimal number of epochs. The evaluation is performed on a k-fold cross-validation strategy and leave one subject out (LOSO) strategy to prove the robustness of the model. For the K-fold CV, each subject along with their EEG trials are randomly partitioned into $k$ folds ($k=5$ for EEGMMIDB; $k=3$ for BCI-IV-2a). In each iteration, the $k-1$ number of folds is used for training and the remaining is used for testing. On the other hand, for LOSO, the model iterates equal to the number of subjects where each subject is set to test while all other subjects work as a train.

\subsection{Results on Different Datasets}
By evaluating the model in both the above mentioned dataset, we show the effectiveness of the SSTAF Transformer architecture. For the EEGMMIDB dataset, the SSTAF Transformer model achieved remarkable results with an average accuracy of 76.83\%, and for BCI Competition IV-2a, the average accuracy is 68.30\%, respectively. To demonstrate the generalization capability of our model, we evaluated in a subject-independent (SI) manner, where mutually exclusive subjects work for training and testing. Given that if the classification algorithm produces promising results in this SI test, it is likely that a new subject can be accommodated by the same model without further training. Our model demonstrated strong performance in this evaluation strategy and generates scores that are compatible with other methods. Table ~\ref{tab:performance} shows the accuracy and F1 scores of the SSTAF Transformer model in the above mentioned datasets for the SI task. In particular, the proposed transformer-based model showed less variability in performance across different subjects, which indicates the adaptability and robustness of our model.

\begin{table}[!ht]
\centering
\caption{Accuracy and F1-score of the proposed model on two benchmark datasets.\label{tab:performance}}
\begin{tabular}{l l c c}
\midrule
\textbf{Dataset} & \textbf{Evaluation Type} & \textbf{Accuracy (\%)} & \textbf{F1-score (\%)} \\
\midrule
EEGMMIDB & Subject-independent & 76.83 & 73.52 \\
\midrule
BCI IV-2a & Subject-independent & 68.30  & 70.63 \\
\midrule
\end{tabular} \\
\end{table}

\subsection{Comparative Analysis with Baseline Models}
We show the classification performance of the SSTAF transformer model in comparison with previous works in Tables ~\ref{tab:cross_subject} and ~\ref{tab:cross_subject2}. Our method outperformed the models in both datasets. Our model not only surpasses traditional CNN-based models such as EEGNet and ShallowConvNet known for their spatial-temporal feature extraction capabilities, but also few existing transformer-based models \cite{tao2021gated} \cite{xie2022transformer} \cite{Zhao2024ctnet} \cite{zhang2019convolutional} or even aligns closely with an advanced transformer-based architecture such as hierarchical transformers \cite{deny2023hierarchical} that emphasize the competitive advantage of transformer architectures in the capture of global dependencies in EEG signals. We mainly focus on the evaluation performances in a cross-subject manner, since this criteria plays an important role in explaining the true nature of the model. In Table ~\ref{tab:cross_subject} we have demonstrated the performance comparison with the transformer-based architecture and in Table ~\ref{tab:cross_subject2} we have compared with baseline architectures based on non-transformers but other representative deep learning-based architectures~\cite{lawhern2018eegnet} \cite{schirrmeister2017deep} \cite{tayeb2019validating} \cite{zhang2018cascade} \cite{roots2020fusion} \cite{zhang2019convolutional}. The results illustrate the reliability of the SSTAF transformer model in the classification of EEG signals, which effectively addressed the inherent variability and complexity of EEG signals.

\begin{table}[!ht]
\centering
\caption{Comparison with the transformer-based models on two datasets for cross-subject evaluation.}
\label{tab:cross_subject}
\renewcommand{\arraystretch}{1.2}
\setlength{\tabcolsep}{6pt}
\begin{tabular}{l l c c}
\toprule
\textbf{Dataset} & \textbf{Model} & \textbf{Subjects} & \textbf{Accuracy (\%)} \\
\midrule
\multirow{3}{*}{EEGMMIDB} 
& GRUGate-Trans~\cite{tao2021gated} & 109 & 61.9 \\
& t-CTrans~\cite{xie2022transformer} & 105 & 68.5 \\
& \textbf{Proposed} & \textbf{103} & \textbf{76.8} \\
\midrule
\multirow{4}{*}{BCI-IV-2a} 
& CTNet~\cite{Zhao2024ctnet} & 9 & 58.6 \\
& TransEEGNet~\cite{zhang2019convolutional} & 9 & 63.5 \\
& Hierarchical-Trans~\cite{deny2023hierarchical} & 9 & 70.3 \\
& \textbf{Proposed} & \textbf{9} & \textbf{68.3} \\
\bottomrule
\end{tabular}
\vspace{0.2cm}
\end{table}

\begin{table}[!ht]
\centering
\renewcommand{\arraystretch}{1.2}
\caption{Classification performance of representative deep learning-based MI-EEG models on EEGMMIDB and BCI Competition IV-2a datasets.\label{tab:cross_subject2}}
\label{tab:combined}
\begin{tabular}{l c c c c}
\toprule
\multirow{2}{*}{\textbf{Model}} & \multicolumn{2}{c}{\textbf{EEGMMIDB}} & \multicolumn{2}{c}{\textbf{BCI IV 2a}} \\
\cmidrule(lr){2-3}\cmidrule(lr){4-5}
& Acc (\%) & AUC & Acc (\%) & AUC \\
\midrule  
EEGNet~\cite{lawhern2018eegnet} & 69.94 & 0.773 & 51.30 & 0.772 \\
Shallow ConvNet~\cite{schirrmeister2017deep} & 75.33 & 0.838 & 56.29 & 0.792 \\
pCNN~\cite{tayeb2019validating} & 57.60 & 0.613 & 36.84 & 0.625 \\
Cascade Model~\cite{zhang2018cascade} & 58.62 & 0.631 & 34.31 & 0.589 \\
Parallel Model~\cite{zhang2018cascade} & 56.73 & 0.591 & 35.42 & 0.621 \\
LSTM Model~\cite{tayeb2019validating} & 49.93 & 0.507 & 26.08 & 0.506 \\
EEGNet Fusion~\cite{roots2020fusion} & 76.18 & 0.849 & 44.31 & 0.759 \\
MBEEGNet~\cite{altuwaijri2022multibranch} & 75.99 & 0.853 & 41.07 & 0.760 \\ 
TS-SEFFNet~\cite{li2021temporal} & 67.08 & 0.759 & 42.59 & 0.740 \\
CTCNN~\cite{schirrmeister2017deep} & 71.28 & 0.754 & 47.67 & 0.770 \\
CRAM~\cite{zhang2019convolutional} & 72.91 & 0.800 & 59.10 & 0.818 \\
\midrule
\textbf{Proposed} & \textbf{76.83} & \textbf{0.810} & \textbf{68.30}  & \textbf{0.781} \\
\bottomrule
\end{tabular}
\end{table}

\subsection{Ablation Study}
To better understand the contribution of each component of our proposed architecture, we conducted a comprehensive ablation study of our model on EEGMMIDB dataset. This analysis evaluates the impact of three key components, such as the spectral attention module, the spatial attention module, and the transformer encoder. By selectively removing each component while keeping others intact, we evaluate their individual contributions to the model's overall performance. We constructed three variants of our baseline SSTAF Transformer model, these are no spectral attention, no partial attention, and no transformer but the hyperparameters for training are identical. The results of the ablation study have been given in Table ~\ref{tab:ablation}.

\begin{table}[!ht]
\centering
\caption{Ablation Study on Model Performance \label{tab:ablation}}
\renewcommand{\arraystretch}{1.5}
\begin{tabular}{l c c}
\toprule
\textbf{Model Variant} & \textbf{Accuracy (\%)} & \textbf{F1-score} \\
\midrule
Full Model & 76.83 & 0.744 \\
No Spectral Attention & 70.21 & 0.688 \\
No Spatial Attention & 72.96 & 0.713 \\
No Transformer & 63.47 & 0.608 \\
\bottomrule \\
\end{tabular}
\end{table}

The spectral attention module allows the model to dynamically emphasize relevant frequency bands that contain discriminative information for MI classification. However, removing the spectral attention module resulted in a 6.62\% decrease in classification accuracy and also a reduction in the F1 score. This performance drop clearly shows the importance of frequency-specific feature weighting in EEG signal analysis. Furthermore, the spatial attention module helps the model focus on the locations of the electrode channels that capture the most discriminative neural activity patterns. The removal of this module led to a decrease in accuracy and F1 score of approximately 3-4\%. Although less prominent than the spectral attention module, the spatial attention module still plays an important role in pointing out the task-relevant EEG channels. In particular, the greatest performance degradation is observed after removing the transformer encoder module, when we observe a reduction of approximately 14\% in both accuracy and F1 score. This substantial decline occurred because of the lack of the critical role of the transformer in capturing complex temporal dependencies and long-range interactions in EEG signals. The multi-attention mechanism within the transformer effectively extracts the critical relationships between time points and frequency-spatial features which helps in better distinction of the brain signals.\\
Therefore, the transformer encoder contributes most significantly to model performance, followed by the spectral attention module and the spatial attention module. These findings validate our architectural design choices and demonstrate that each component plays an important role in addressing the unique characteristics of EEG signals.

\subsection{Visualization Experiment}
This section explains the visualization experiment, including the visualization of the topological map, feature maps, and attention weights.

\subsubsection{Topographical Map}
\begin{figure}[!h]
    \centering
    \includegraphics[width=0.5\textwidth]{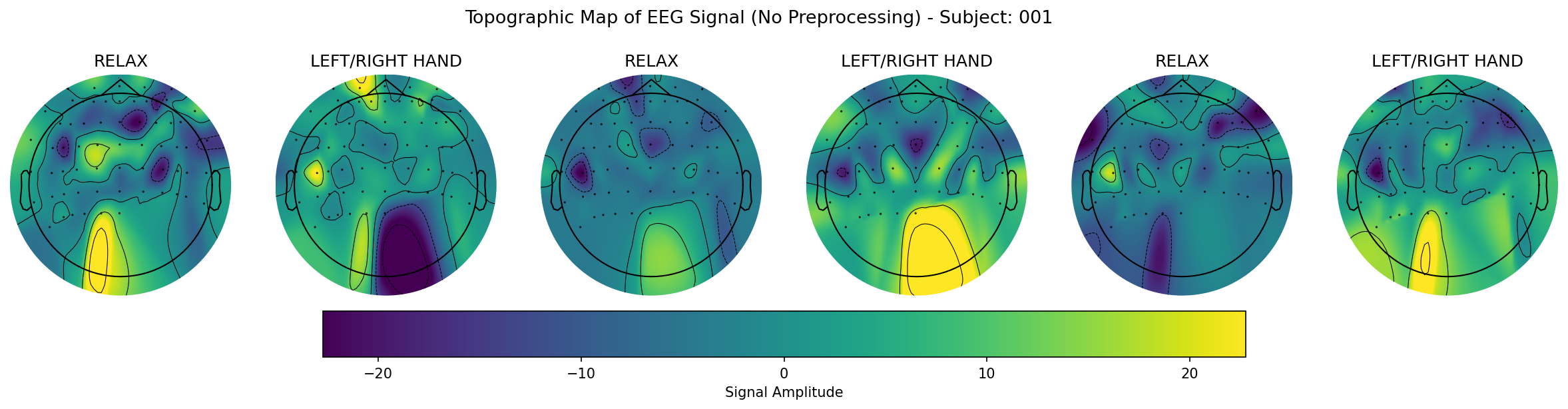}
    \caption{\textbf{Topographical Map}, Visualization of the raw EEG signals for left and right-hand MI tasks of a single subject (001) and a single run.\label{fig:topographical_map_base}}
\end{figure}

\begin{figure}[!h]
    \centering
    \includegraphics[width=0.5\textwidth]{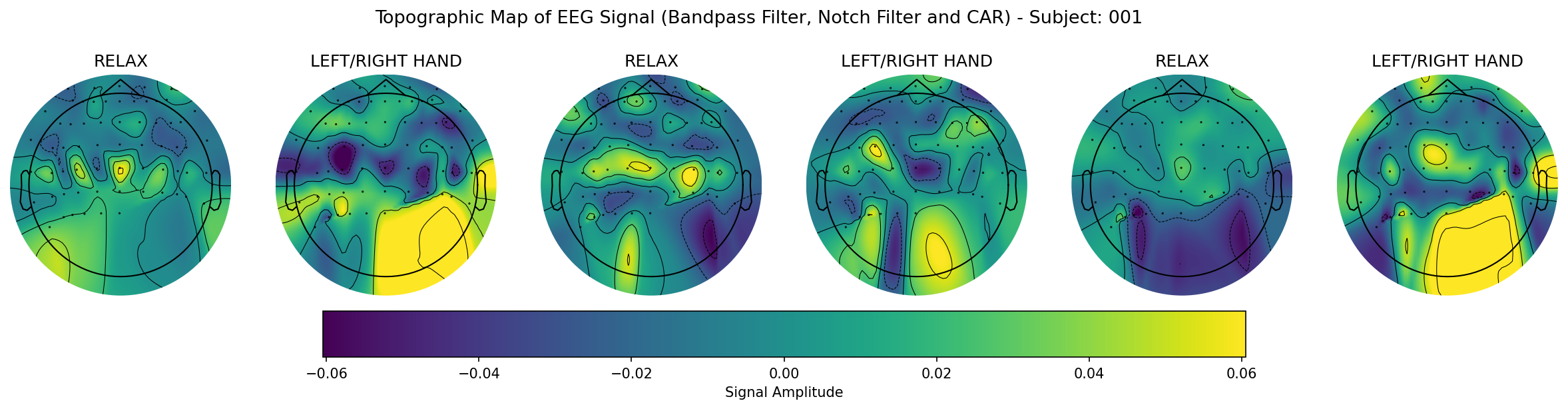}
    \caption{\textbf{Topographical Map after Preprocessing}, Visualization of the EEG signals for left and right-hand MI tasks after preprocessing of the same subject and the same run as shown in Figure~\ref{fig:topographical_map_base}\label{fig:topographical_map_preprocessed}.}
\end{figure}

To visualize the spatial distribution of EEG signals between different electrodes, we generate a topographic map of before and after preprocessing the raw EEG signals. We illustrate the maps for MI tasks of relax and hand movements of a randomly chosen subject for the EEGMMIDB dataset in Figure~\ref{fig:topographical_map_base} and Figure~\ref{fig:topographical_map_preprocessed} respectively. For a particular run, the first six segments of the EEG signals are selected. The topographical map is generated for each channel-wise averaged segment and is color-coded to represent the amplitude of the EEG signals. It is clearly observed that after preprocessing, certain regions of the brain exhibited significant changes in the distribution of EEG signals, specifically for hand movements. These changes make the signals better distinguishable, helping in later classification stages.  

\subsubsection{STFT Feature Map}
\begin{figure}[!ht]
    \centering
    \includegraphics[width=0.5\textwidth]{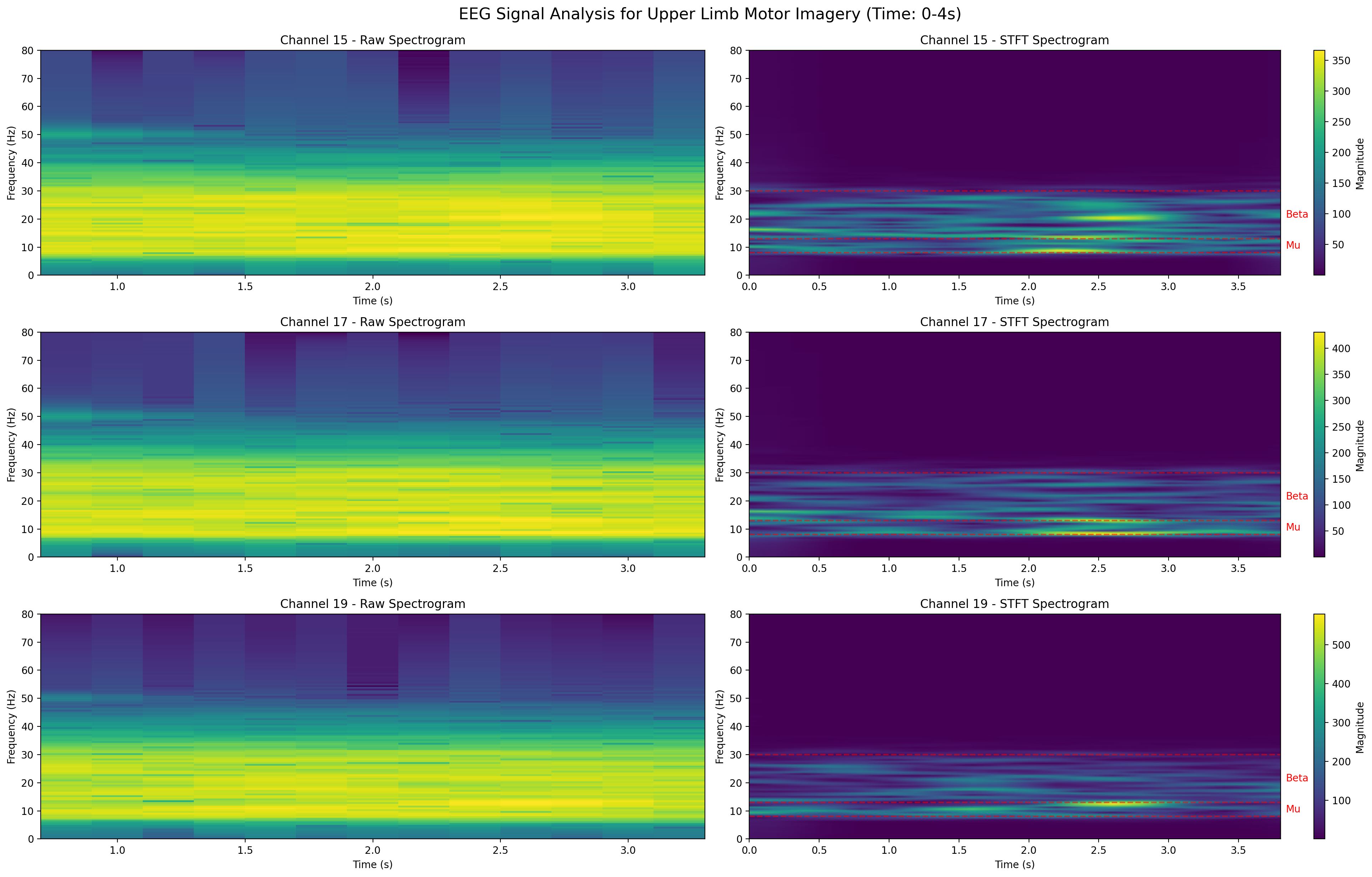}
    \caption{\textbf{STFT Feature Map}, Visualization of the raw and STFT feature map using spectrogram for left and right-hand MI tasks.\label{fig:stft_feature_map}}
\end{figure}

The STFT feature map provides information on the frequency content of EEG signals over time. We generate an STFT feature map by applying the STFT to the EEG signals after preprocessing to get the time-frequency representation of the data. In Figure~\ref{fig:stft_feature_map}, the raw EEG signal and the STFT feature map are compared using the spectrogram visualization on the left and right sides of the figure for channels 15, 17, and 19. In this study, we focus on the band pass frequency range of 8 Hz to 30 Hz, which is commonly associated with MI tasks. Therefore, in the STFT feature map, the $\mu$ and $\beta$ bands are highlighted and enhanced. This STFT input representation enhances the model's ability to discern patterns and improve classification accuracy.

\subsubsection{Attention Weights Visualization}
\begin{figure}[!h]
\centering
    \includegraphics[width=0.48\textwidth]{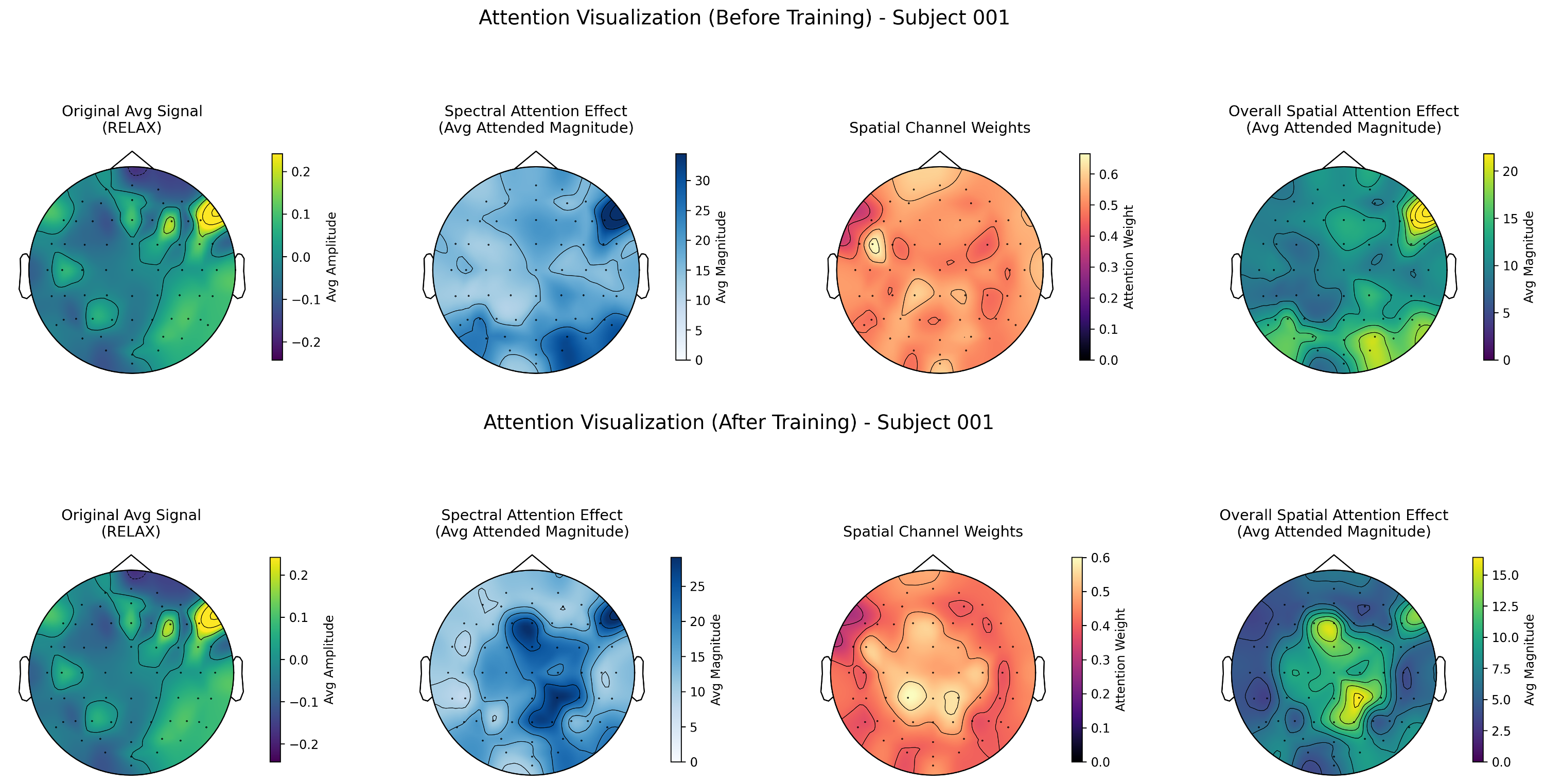}
    \caption{\textbf{Comparison of relaxed state attention heatmap}, first row represents untrained model and the second row represents the trained model.\label{fig:relax_comparison_model}}
\end{figure}

\begin{figure}[!h]
\centering
    \includegraphics[width=0.48\textwidth]{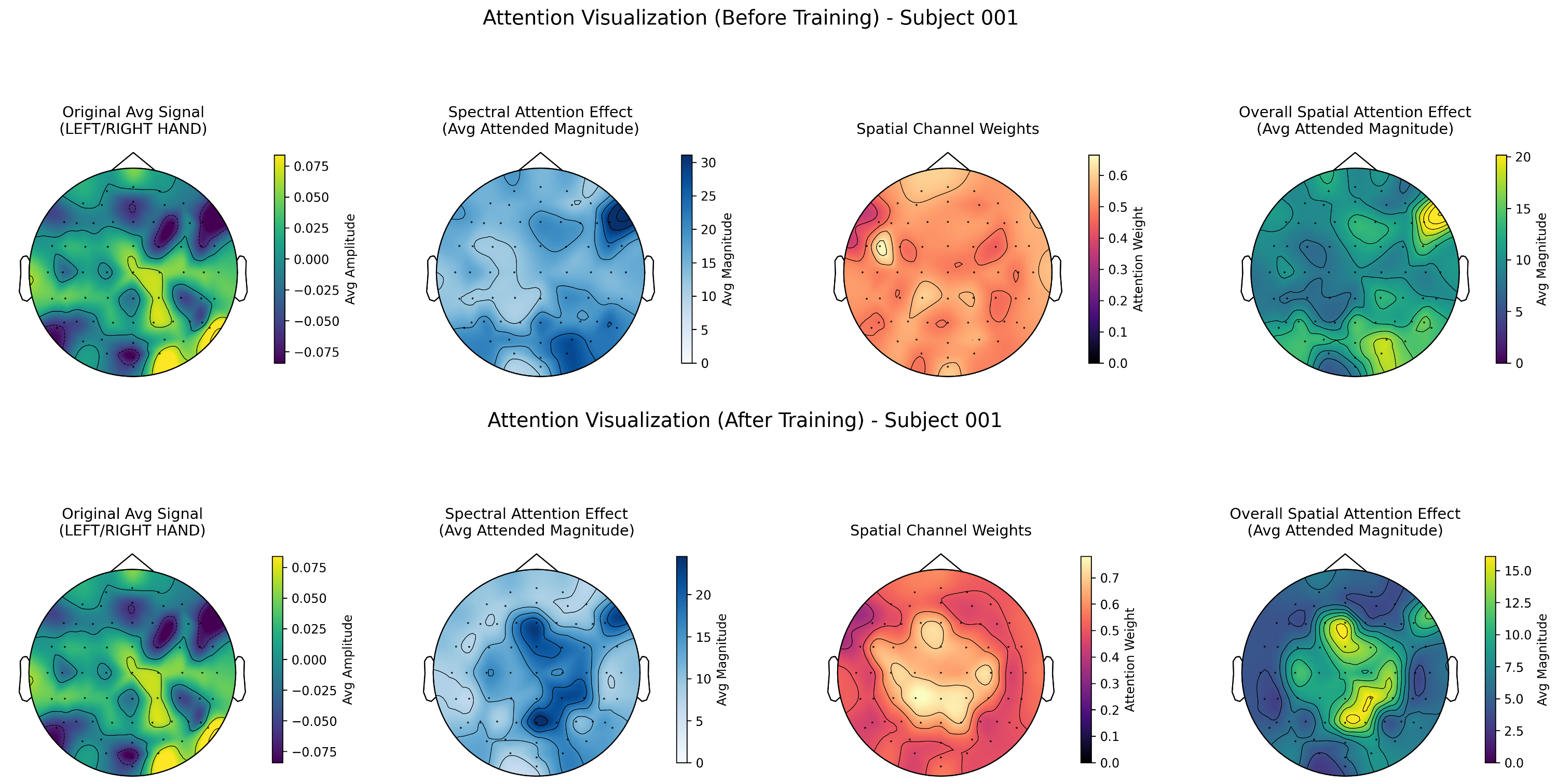}
    \caption{\textbf{Comparison of hand movement state attention heatmap}, first row represents untrained model and the second row represents the trained model.\label{fig:hand_comparison_model}}
\end{figure}

Figure~\ref{fig:relax_comparison_model} and~\ref{fig:hand_comparison_model} compare the difference in feature highlighting capabilities between the untrained and trained models for both the relax state and the left/right-hand MI condition. Each row comprises four topographic maps that show the raw average EEG signal, the spectral attention effect, the spatial channel weights, and the overall attention effect, respectively. Relaxation state and hand movement data from the first subject were selected for this visualization.

Before training the model, it is seen that for both the relax and the movement cases of the hand, the spectral and spatial attention module of the model applies random attention to the EEG signals. In the spectral attention module, the critical channels are not highlighted as expected, and it has not learned to focus on the $\mu$ and $\beta$ bands (8--30 Hz) for MI movement. Similarly, for the spatial attention module, the attention weights of the expected motor region areas are not highlighted yet.

However, after training the model, the attention heatmaps show a significant improvement in the model's ability to focus on the relevant frequency bands and spatial channels. The spectral attention module now highlights the $\mu$ and $\beta$ bands, although the visualizations are in terms of spatial perspective rather than signal frequency. In the relax state, the motor cortex regions show synchronous $\mu$ and $\beta$ bands that tend to increase while relaxing, and the spectral attention module highlighted the expected regions. Moreover, in the hand movement state, some expected channels such as \texttt{C3, C4, Cz} show significant activation patterns, and the model's heatmap shows high attention in those areas. As for spatial attention weight, for both cases it is seen that the model learned to attend more to the sensorimotor cortex region. 

In summary, the training process significantly refines the network's attention mechanisms. It is clearly visible that the model increasingly focuses on the sensorimotor regions and demonstrates enhanced differentiation of EEG features as it is trained.

\subsection{Key Findings}
Our model incorporates three key components addressing the cross-subject variability challenges. The dual attention mechanism helps the model capture discriminative EEG patterns. The spectral attention module enhanced classification accuracy due to its selective emphasis on discriminative spectral features, which helps overcome individual differences in frequency-domain characteristics of MI patterns. Despite variations in exact spatial patterns between subjects, the spatial attention module could identify electrode locations corresponding to motor cortex regions associated with different MI tasks. Finally, the Transformer Encoder's self-attention mechanism effectively captures long-range dependencies and contextual relationships in EEG signals, allowing the model to learn more robust temporal representations that generalize across subjects. Additionally, implementing a rigorous cross-subject with cross-validation methodology helps to address the generalization problem by ensuring that the model never sees test subjects during training. The model can better distinguish EEG patterns by collectively attending to all complementary domains.

Lastly, the performance of our model could not increase more, primarily due to the challenges inherent in the dataset itself. Inherent EEG variability, non-stationarity signals, and limited training data likely constrain it. EEG data contains low SNR, where artifacts and background activity often overwhelm neural signals of interest. Moreover, significant inter-subject variability and different spatial organization between subjects also hinder the performance. Another crucial aspect behind the bounded performance of the model is the limited dataset size, which is specifically problematic for transformer-based models. On the other hand, to make the model less complex, we use a relatively small transformer, which may limit its capacity to learn complex patterns. In the classifier module, we use mean pooling across time frames that also may lose important temporal dynamics in the signal.

\section{Conclusion}
This study introduced the SSTAF Transformer model for EEG-based MI classification, demonstrating a few crucial aspects necessary for EEG signals in terms of deep learning models. Using a novel attention fusion mechanism that integrates information across the spectral, spatial, and temporal domains of EEG signals, our model effectively captures the complex cross-dimensional interactions essential for accurate MI classification and outperforms several existing approaches. The extraction of STFT features allows the SSTAF Transformer model to have meaningful features, helping to better spectral-spatial transformation. Our cross-subject validation strategy ensures the appropriate separation between training and testing data, producing unbiased results from the model. Finally, our model demonstrates superior performance across benchmark datasets, including the EEGMMIDB and BCI Competition IV-2a.  Overall, the SSTAF Transformer contributes to EEG-based MI classification by highlighting an effective strategy that includes multi-domain aspects of the EEG signals. Our future work will focus on collecting a large, high-quality dataset with a high SNR to enhance cross-subject generalization. We also aim to explore adaptive personalization mechanisms for real-time implementation, subject-specific calibration, and improved temporal feature integration to further enhance classification performance.

\footnotesize
\bibliographystyle{IEEEtran}
\bibliography{references}

\end{document}